%% file: neurips.tex
\documentclass{article}

\usepackage[main, final]{neurips_2026}

\usepackage[utf8]{inputenc}
\usepackage[T1]{fontenc}
\usepackage{hyperref}
\usepackage{url}
\usepackage{booktabs}
\usepackage{amsfonts}
\usepackage{nicefrac}
\usepackage{xspace}
\usepackage{microtype}
\usepackage{xcolor}
\usepackage{graphicx}
\usepackage{subcaption}

\usepackage{amsmath}
\usepackage{amssymb}
\usepackage{algorithm}
\usepackage{algpseudocode}
\usepackage{multirow}
\usepackage{enumitem}
\usepackage{listings}
\usepackage{wrapfig}

\newcommand{\myparatight}[1]{\smallskip\noindent{\bf {#1}:}~}
\newcommand{\method}{RuleEvolve\xspace}
\DeclareMathOperator*{\argmax}{arg\,max}

\title{Self-Evolving Coding Rules for AI Coding Agents}

\author{
  Zhengyuan Jiang \\
  Duke University \\
  \texttt{zhengyuan.jiang@duke.edu} \\
  \And
  Reachal Wang \\
  Duke University \\
  \texttt{reachal.wang@duke.edu} \\
  \And
  Yuepeng Hu \\
  Duke University \\
  \texttt{yuepeng.hu@duke.edu} \\
  \And
  Yupu Wang \\
  Duke University \\
  \texttt{yupu.wang@duke.edu} \\
  \And
  Yuqi Jia \\
  Duke University \\
  \texttt{yuqi.jia@duke.edu} \\
  \And
  Neil Zhenqiang Gong \\
  Duke University \\
  \texttt{neil.gong@duke.edu} \\
}

\begin{document}
\maketitle

\input{0_abstract}

\input{1_intro}
\input{2_background}
\input{3_problem}
\input{4_method}
\input{5_evaluation}
\input{6_conclusion}

\bibliography{main}
\bibliographystyle{plain}

\appendix
\input{8_appendix}


\end{document}

%% file: 0_abstract.tex
\begin{abstract}
The performance of AI coding agents is highly dependent on their underlying coding rules. However, existing coding rules are typically hand-crafted and fixed, making the process labor-intensive and often suboptimal. In this work, we propose \emph{\method}, a self-evolving framework for coding rules. \method maintains a pool of candidate coding rules and iteratively improves them. In each iteration, it employs an LLM-powered \emph{mutator module} to generate variants from existing candidates, and then uses a \emph{judge module} to evaluate these variants and update the pool with the best-performing ones. Extensive evaluations across two coding-agent frameworks, four backbone LLMs, and three benchmarks demonstrate that \method outperforms both manual engineering and existing prompt optimization baselines in terms of functional correctness of the generated code, code length, and/or generation cost (e.g., tokens used).
\end{abstract}

%% file: 1_intro.tex
\section{Introduction}
The evolution of AI coding agents, such as OpenHands~\cite{wangopenhands}, OpenCode~\cite{opencode}, Claude Code~\cite{claudecode}, and Cursor~\cite{cursor}, has fundamentally redefined code generation tasks by shifting the focus from manual implementation to high-level orchestration. Central to the performance of these agents is the coding rules file, often instantiated as an \texttt{AGENTS.md} document and prepended to the backbone model's input context~\cite{agentsmd}. These rules play a critical role in determining the correctness and length of the generated code, as well as the associated generation cost.

Despite the critical importance of these rules, they are currently almost exclusively hand-crafted by human developers~\cite{githubrules}. This manual process is inherently labor-intensive and suboptimal, as human-written rules often lack task-specific alignment and can even degrade agent performance~\cite{gloaguen2026evaluating}.

Although automatic techniques~\cite{wu2025llm} for prompt optimization have been developed, these methods remain difficult to directly apply to self-evolving coding rules. Existing approaches primarily target short, few-sentence instructions, whereas coding rules are typically long, well-documented guidance that present a significantly larger search space. Furthermore, the resource-intensive nature of coding agents renders previous random mutation strategies computationally prohibitive. Consequently, there is a critical need for a framework that can automatically evolve complex coding rules.

\begin{figure*}[h]
    \centering
    \includegraphics[width=\linewidth]{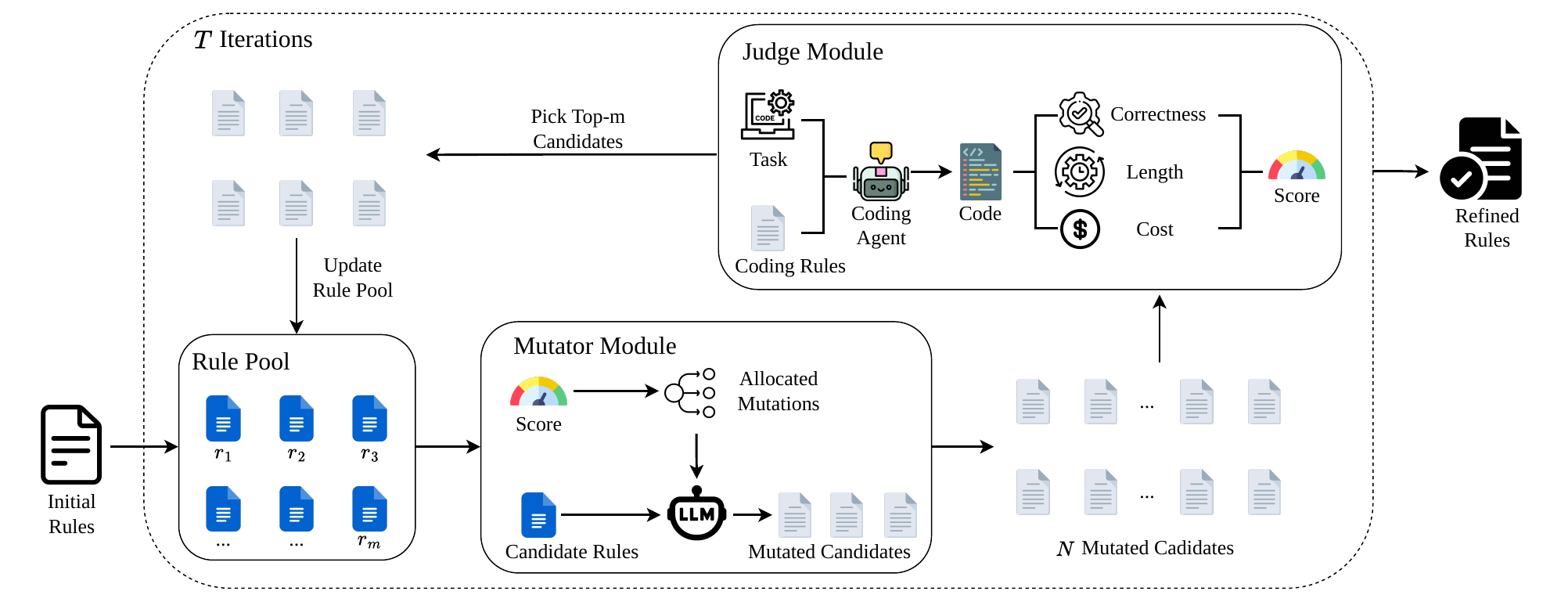} 
    \caption{Pipeline of \method.}
    \label{fig:pipeline}
\end{figure*}

To bridge this gap, we propose \method, a self-evolving framework for coding rules. The overall pipeline is illustrated in Figure~\ref{fig:pipeline}. \method takes as input a set of coding tasks and aims to evolve coding rules that can generalize to improve code generation on other tasks. Specifically, \method maintains a pool of $m$ candidate coding rules and iteratively evolves them. In each iteration, it employs an LLM-powered \emph{mutator module} to generate $N$ mutations from the current candidates. Each mutation is then evaluated by a \emph{judge module}, which assesses performance in terms of code correctness, code length, and generation cost. Based on these evaluations, \method updates the pool by retaining the top-performing $m$ candidates.

One challenge is how to allocate the $N$ mutations among the $m$ candidates in the pool in each iteration.  Naive strategies, which distribute mutations equally or proportionally based on each candidate's performance, produce suboptimal results, as demonstrated in our experiments. To enhance effectiveness, we propose a strategic mutation-allocation method, which dynamically distributes the $N$ mutations by prioritizing candidates with higher performance variance or those whose mean performance is closer to the current best.

We evaluate \method across two coding-agent frameworks, four backbone LLMs, and three coding benchmarks. Extensive experiments show that the proposed method consistently maintains or improves functional correctness measured by pass rate, while significantly reducing code length and generation cost, as measured by generation time and token usage. For instance, on the BigCodeBench dataset using the OpenAI SDK with gpt-5.3-codex, \method successfully reduced average code length by over 60\% without sacrificing pass rate.

%% file: 2_background.tex
\section{Related Works}

\subsection{AI-assisted Code Generation}
The landscape of software engineering has been transformed by AI-assisted code generation. To evaluate these AI-generated code, the research community has established several coding benchmarks, such as HumanEval~\cite{chen2021evaluating} and BigCodeBench~\cite{zhuobigcodebench}, which assess the ability of models to generate executable and functional code across a variety of programming tasks. Recent works~\cite{midolo2026guidelines,ye2025prompt} have sought to enhance the quality of generated code through prompting strategies. However, these efforts primarily focus on the performance of the underlying coding LLM in isolation rather than within the agentic framework. Furthermore, these studies focus on the optimization of task-specific prompts to the coding LLM, without considering the self-evolving of the coding rules that guide the code generation process universally.

\subsection{Coding Rules for AI Coding Agents}
AI coding agents rely on coding rules, typically stored in an \texttt{AGENTS.md} file, to guide the generation of code. These rules and guidance are combined with the system and user prompts to serve as input for the backbone LLM. The specific rules provided influence the structure and style of the generated code. While recent works~\cite{galster2026configuring,gloaguen2026evaluating,chen2026dynamic} have explored the impact of these rules on code generation, these studies primarily focus on measurement without providing practical insights into how these coding rules can be systematically improved or evolved.

\subsection{Self-evolving Agents}
Existing studies on self-evolving agents primarily focus on models, tools, skills, and memory~\cite{yang2026skillopt,shao2026your,sun2025seagent,zhang2026coevoskills}. While coding rules are central to coding agents, their self-evolving remains largely unexplored. Automatic prompt optimization techniques~\cite{wu2025llm,khattab2024dspy,tong2025evoprompt,fernando2023promptbreeder,nag2025mutantprompt,mehrotra2024tree,chao2025jailbreaking,jiang2026jailbreaking} can theoretically be applied to optimize coding rules; however, these methods primarily target short prompts consisting of only a few sentences. In contrast, coding rules are extensive, well-documented sets of instructions that create a significantly larger search space beyond the capability of existing optimizers. Consequently, applying current prompt optimization techniques directly to coding rules yields suboptimal performance, as confirmed by our experiments with Prompt-Ops~\cite{wu2025llm}.

%% file: 3_problem.tex
\section{Problem Definition}

\subsection{AI Agents for Code Generation}
We consider an AI coding agent $\mathcal{A}_\theta$ powered by a backbone  LLM with parameters $\theta$. The agent generates code conditioned on a specific programming task and underlying coding rules. In practice, for most modern coding agents, these rules are instantiated via an \texttt{AGENTS.md} document and prepended to the LLM's input context.

Let $x \in \mathcal{X}$ denote a coding task (i.e., the problem description), $r$ denote the coding rules, and $y$ denote the generated code, where $\mathcal{X}$ is the coding task set. Formally, the coding agent $\mathcal{A}_\theta$ samples the code $y$ from the conditional distribution:
\begin{align}
    y \sim p_\theta(\cdot \mid x, r).
\end{align}
Given a coding task $x$ and fixed parameters $\theta$, the coding rules $r$ significantly influence the code generation process, thereby affecting the correctness, length, and inference cost. While correctness ensures the functional integrity of the code, length and inference cost are equally critical for real-world deployment. Code length determines the performance of the generated code, where optimized logic reduces execution latency and resource consumption. Inference cost directly affect the scalability of the agentic system; by minimizing token usage and generation time, developers can substantially lower API expenses and improve the responsiveness of the development cycle. Formally, we define two primary goals for the coding rule self-evolving process:

\begin{itemize}
    \item The generated code $y$ should be executable and functionally correct.
    
    \item The evolved coding rules should reduce the code length as possible and minimize inference costs during the generation process.
\end{itemize}

\subsection{\label{sec:objective}Objective}
To satisfy the first goal, we define a correctness constraint. Let $\mathcal{V}(x, y) \in \{0,1\}$ be a code verifier that determines whether code $y$ is executable and correctly solves the programming task $x$, where $\mathcal{V}(x, y)=1$ indicates success. In practice, $\mathcal{V}$ may be implemented by executing $y$ against a set of test cases associated with $x$. The evolved coding rules $r$ must satisfy:
\begin{align}
    \frac{1}{|\mathcal{X}|} \sum_{x \in \mathcal{X}} \mathbb{E}_{y \sim p_\theta(\cdot \mid x, r)} [\mathcal{V}(x, y)] \geq \tau,
\end{align}
where $\tau$ is a predefined threshold ensuring that the average correctness rate of the agent $\mathcal{A}_\theta$ across the task set $\mathcal{X}$ remains acceptable.

To satisfy the second goal, we define a code-length score $E(y)$ and a cost score $C(\mathcal{A}_\theta, x, r)$. A higher $E(y)$ indicates that the generated code $y$ is more concise, while a higher $C(\mathcal{A}_\theta, x, r)$ indicates lower resource consumption during the generation process. Formally, the coding rule self-evolving process is expressed as the following constrained optimization problem:
\begin{align}
    \max_{r} \quad & \frac{1}{|\mathcal{X}|} \sum_{x \in \mathcal{X}} \mathbb{E}_{y \sim p_\theta(\cdot \mid x, r)} [E(y) + C(\mathcal{A}_\theta, x, r)] \nonumber \\
    \text{s.t.} \quad & \frac{1}{|\mathcal{X}|} \sum_{x \in \mathcal{X}} \mathbb{E}_{y \sim p_\theta(\cdot \mid x, r)} [\mathcal{V}(x, y)] \geq \tau.
\end{align}

%% file: 4_method.tex
\section{\method}

\subsection{Overview}
To achieve these objectives, we propose \method, a self-evolving framework for coding rules. As illustrated in Figure~\ref{fig:pipeline}, \method consists of three primary components: a rule pool of fixed size $m$, a \emph{mutator module} for generating mutated candidates, and a \emph{judge module} for evaluation and selection. Given the initial coding rules $r$, \method first employs the mutator module to populate the pool with $m$ mutated versions. In each subsequent iteration, $N$ newly mutated candidates are generated from $m$ candidates in the current pool. Mutations are strategically allocated based on the historical mean and variance of the candidates' scores. Finally, the $m+N$ candidates are evaluated by the judge module; the top $m$ candidates are retained for the next iteration, while others are pruned.

\myparatight{Challenge and motivation} The self-evolving of coding rules presents two key challenges. First, the generation process in a coding agent typically involves multi-turn interactions and external tool-calling, rendering the pipeline non-differentiable and precluding direct gradient-based optimization. We address this by implementing an iterative mutation and pruning framework that leverages the mutator module and the judge module to approximate the optimization process (see Section~\ref{sec:design}). Second, unlike standard prompt optimization tasks that involve short targets and low evaluation costs, coding rules are significantly more resource-intensive to evolve. Each rule set requires full agent execution and verification, leading to substantial latency and API token consumption. To avoid resource waste and enhance effectiveness, we propose a strategic mutation-allocation method that prioritizes mutating promising candidates, as detailed in Section~\ref{sec:mutation allocation}.

\subsection{\label{sec:design}Iterative Mutation and Pruning}

\myparatight{Mutator $\mathcal{M}$} The mutator module, powered by an LLM, takes the coding rules $r$ and a mutation guidance $g$ as input to produce a mutated candidate $r_m = \mathcal{M}(r, g)$. The mutation guidance $g$ typically consists of a natural language directive, such as: \emph{"Mutate the coding rules for an AI coding agent into a new variant by improving wording, reordering or combining steps, and varying the style."}

\myparatight{Judge $\mathcal{J}$} The Judge module provides the evaluation environment for generating code under specific coding rules, executing the generated code, and assessing performance across the dimensions of correctness, length, and inference cost. Specifically, the Judge $\mathcal{J}$ takes the coding rules $r$, the agent $\mathcal{A}_\theta$, and a task set $\mathcal{X}$ as input to produce a score $S = \mathcal{J}(r, \mathcal{A}_\theta, \mathcal{X})$.

As formulated in Section~\ref{sec:objective}, this score consists of two components. The first is a correctness penalty designed to ensure that the evolved rules do not degrade the agent's functional performance. Given the code verifier $\mathcal{V}$, the penalty is defined as:
\begin{align}
    \mathbb{E}_{x \in \mathcal{X}} \left[ \mathcal{V}(x, \mathcal{A}_\theta(x, \emptyset)) - \mathcal{V}(x, \mathcal{A}_\theta(x, r)) \right]_+,
\end{align}
where $\mathcal{A}_\theta(x, \emptyset)$ denotes the code generated by the agent without external coding rules, $\mathcal{A}_\theta(x, r)$ denotes the code generated using coding rules $r$, and $[z]_+ = \max(z, 0)$.

The second component focuses on the code length and inference cost of the generation process. As illustrated in Figure~\ref{fig:iterations}, there is a high correlation between code characters and token usage during generation. Consequently, we approximate both code length and inference cost using the number of characters of the generated code, denoted by $|\mathcal{A}_\theta(x, r)|$. The overall score $S$ is formulated as:
\begin{align}
    S = -|\mathcal{A}_\theta(x, r)| - \lambda \cdot \mathbb{E}_{x \in \mathcal{X}} \left[ \mathcal{V}(x, \mathcal{A}_\theta(x, \emptyset)) - \mathcal{V}(x, \mathcal{A}_\theta(x, r)) \right]_+,
\end{align}
where $\lambda$ is a hyperparameter that balances the trade-off between two components.

\myparatight{Iterative Process} The complete procedure is summarized in Algorithm~\ref{alg:evolving}. Our approach begins by populating an initial candidate rule pool $\mathcal{P}$ through $m$ mutations of the initial coding rules. In each iteration $t$, we strategically distribute $N$ mutations among existing $m$ candidates based on their historical scores. This results in $N$ newly mutated candidates by the mutator module $\mathcal{M}$, which are subsequently evaluated by the judge module $\mathcal{J}$. To maintain a manageable search space and focus on high-performing regions, we employ a pruning step that retains only the top-$m$ candidates for the subsequent iteration. After $T$ iterations, the framework returns the coding rules that achieved the highest average score.

\begin{algorithm*}[t]
\caption{\method: Coding Rule Self-Evolving}
\label{alg:evolving}
\begin{algorithmic}[1]
    \Require Initial rules $r_I$, task set $\mathcal{X}$, rule pool size $m$, number of mutations $N$, agent $\mathcal{A}_\theta$, mutator $\mathcal{M}$, judge $\mathcal{J}$, max iterations $T$
    \State $\mathcal{P} \gets \mathcal{M}(r_I, m)$ \Comment{Initialize pool with $m$ mutated candidates}
    \State $\mathcal{S} \gets \mathcal{J}(\mathcal{P}, \mathcal{A}_\theta, \mathcal{X})$ \Comment{Obtain initial scores for the pool}
    \For{$t=1$ to $T$}
        \State $\{n_1, \dots, n_m\} \gets \textsc{MutationAllocation}(\mathcal{S}, N)$ \Comment{Allocate $N$ mutations strategically}
        \State $\mathcal{P}_{new}, \mathcal{S}_{new} \gets \emptyset, \emptyset$
        \For{$j=1$ to $m$}
            \State $\mathcal{R}_{new} \gets \mathcal{M}(\mathcal{P}_j, n_j)$ \Comment{Generate $n_j$ new mutations from candidate $j$}
            \State $\mathcal{P}_{new} \gets \mathcal{P}_{new} \cup \mathcal{R}_{new}$
            \State $\mathcal{S}_{new} \gets \mathcal{S}_{new} \cup \{ \mathcal{S}_j \text{ for each } r \in \mathcal{R}_{new} \}$ \Comment{Inherit score set from parent $j$}
        \EndFor
        \State $\mathcal{S}_{eval} \gets \mathcal{J}(\mathcal{P}_{new}, \mathcal{A}_\theta, \mathcal{X})$ \Comment{Evaluate new candidates}
        \State $\mathcal{S}_{new} \gets \text{Update}(\mathcal{S}_{new}, \mathcal{S}_{eval})$ \Comment{Update score sets}
        \State $\mathcal{P}, \mathcal{S} \gets \textsc{Prune}(\mathcal{P} \cup \mathcal{P}_{new}, \mathcal{S} \cup \mathcal{S}_{new}, m)$ \Comment{Retain top-$m$ candidates}
    \EndFor
    \State \Return $\argmax_{r \in \mathcal{P}} \text{avg}(\mathcal{S}_r)$
\end{algorithmic}
\end{algorithm*}

\subsection{\label{sec:mutation allocation}Strategic Mutation Allocation}
To avoid meaningless mutations and enhance effectiveness, we employ the Optimal Computing Budget Allocation~\cite{chen2011stochastic} to distribute $N$ mutations among $m$ candidates wisely. The core objective is to maximize the probability of correctly identifying the best-performing candidate while minimizing redundant evaluations of suboptimal candidates.

For each candidate $i$, we maintain a score set $\mathcal{S}_i$ containing the score from each iteration's evaluation. To address the cold-start problem and ensure statistical stability, a newly mutated candidate inherits the score set from its parent. This inheritance ensures that all candidates in the rule pool possess the same number of observations, allowing for a consistent computation of the mean $\hat{\mu}_i$ and standard deviation $\hat{\sigma}_i$ for candidate $i$.

Let $b$ be the index of the current best-performing candidate in the pool, i.e., with the highest average score. We determine the allocation weights $w_i$ for all $i \neq b$ such that:
\begin{align}
    w_i \propto \left( \frac{\hat{\sigma}_i}{\hat{\mu}_b - \hat{\mu}_i} \right)^2.
\end{align}
This formulation ensures that more mutations are allocated to candidates with higher performance variance or those whose mean performance is closer to the current best. The weight for the best candidate $w_b$ is then calculated as follows to maintain overall allocation efficiency:
\begin{align}
    w_b = \hat{\sigma}_b \sqrt{\sum_{i \neq b} \frac{w_i^2}{\hat{\sigma}_i^2}}.
\end{align}
Finally, the discrete $n_i$ mutations for each candidate is determined by normalizing these weights such that $\sum n_i = N$. By iteratively refreshing these statistics, updating allocation weights $w_i$, and redistributing the mutations $n_i$, \method focuses its search on the most promising regions of the coding rule space, improving the effectiveness of the evolving process. Experimental results validate this by comparing with naive mutation-allocation strategies. More details can be found in Appendix~\ref{app:ocba}.

\begin{algorithm}[t]
\caption{MutationAllocation}
\label{alg:mutation_allocation}
\begin{algorithmic}[1]
    \Require Score sets $\mathcal{S}_i$ for candidates $\{1, \dots, m\}$, number of mutations $N$
    \State Compute means $\hat{\mu}_i$ and standard deviations $\hat{\sigma}_i$ from $\mathcal{S}_i$ for each $i \in \{1, \dots, m\}$
    \State $b \gets \arg\max_i \hat{\mu}_i$
    \For{$i=1$ to $m, i \neq b$}
        \State $\Delta_i \gets \max(\hat{\mu}_b - \hat{\mu}_i, \epsilon)$
        \State $w_i \gets (\hat{\sigma}_i / \Delta_i)^2$
    \EndFor
    \State $w_b \gets \hat{\sigma}_b \sqrt{\sum_{i \neq b} (w_i^2 / \hat{\sigma}_i^2)}$
    \State $\{n_1, \dots, n_m\} \gets \text{round}\left(N \cdot \frac{w_i}{\sum_j w_j}\right)$
    \State \Return $\{n_1, \dots, n_m\}$
\end{algorithmic}
\end{algorithm}

%% file: 5_evaluation.tex
\section{Evaluation}

\subsection{Experimental setup}

\myparatight{Datasets} We utilize the BigCodeBench benchmark~\cite{zhuobigcodebench} as our primary dataset for coding rules evolvement. From this benchmark, we randomly sample 1,000 instances to serve as a training set. During the self-evolving process, each iteration's scores are obtained using mini-batches of size $B$, sampled randomly from this training set. To evaluate the performance of evolved coding rules, we employ three coding benchmarks: BigCodeBench, BigCodeBench-Hard~\cite{zhuobigcodebench}, and HumanEval~\cite{chen2021evaluating}. For each of these benchmarks, we randomly select 100 samples as our testing benchmark. Samples drawn from BigCodeBench for testing are strictly disjoint from the training sets used during evolvement.

\myparatight{Coding agents} We evaluate our method using two distinct agentic frameworks: OpenHands~\cite{wangopenhands}, a state-of-the-art open-source autonomous agent, and a custom GPT-based agent implemented via the OpenAI SDK. For OpenHands, our evolved coding rules are provided through the AGENTS.md configuration, while for OpenAI SDK, rules are integrated directly into the system prompt.

\myparatight{Backbone models} To evaluate the generalizability of coding rules evolvement approaches, we conduct experiments using both proprietary and open-weight models. For the proprietary models, we utilize gpt-4.1-mini, gpt-5-mini, and the state-of-the-art gpt-5.3-codex, all accessed via the OpenAI API. For the open-weight model, we employ Qwen-Coder-30b \cite{qwen3}. We evolve the coding rules for each model respectively.

\begin{table*}[t]
\centering
\caption{OpenAI SDK results. For each method, we report the performance of the backbone that achieved the highest pass rate across benchmarks. Metrics include Pass Rate (\textbf{PR}), Avg. Code Lines (\textbf{CLn}), Avg. Code Characters (\textbf{CC}), Avg. Generation Time (\textbf{Time}), and Avg. Token Usage (\textbf{Tok}).}
\label{tab:openai-sdk-consolidated}
\resizebox{0.99\textwidth}{!}{
\setlength{\tabcolsep}{4pt}
\begin{tabular}{l ccc ccc ccc ccc ccc}
\toprule
\multirow{2}{*}{\textbf{Method}} & \multicolumn{3}{c}{\textbf{PR} $\uparrow$} & \multicolumn{3}{c}{\textbf{CLn} $\downarrow$} & \multicolumn{3}{c}{\textbf{CC} $\downarrow$} & \multicolumn{3}{c}{\textbf{Time} $\downarrow$} & \multicolumn{3}{c}{\textbf{Tok} $\downarrow$} \\
\cmidrule(lr){2-4} \cmidrule(lr){5-7} \cmidrule(lr){8-10} \cmidrule(lr){11-13} \cmidrule(lr){14-16}
& BCB & BCB-H & HE & BCB & BCB-H & HE & BCB & BCB-H & HE & BCB & BCB-H & HE & BCB & BCB-H & HE \\
\midrule
No rule     & 0.47 & 0.25 & 0.97 & 35.7 & 42.3 & 26.4 & 1156 & 1406 & 812.4 & 6.00 & 5.44 & 4.24 & 623.5 & 666.5 & 502.9 \\
Manual rule & 0.45 & 0.23 & 0.98 & 69.0 & 84.8 & 44.4 & 2372 & 2946 & 1136  & 8.75 & 10.5 & 5.08 & 1022 & 1147 & 663.8 \\
Prompt-Ops  & 0.48 & 0.23 & 0.89 & 13.8 & 15.7 & 4.50 & 474.4 & 583.0 & 126.5 & 2.30 & 2.90 & 1.33 & 581.7 & 588.7 & 406.3 \\
\method     & 0.49 & 0.26 & 0.99 & 12.6 & 20.8 & 4.40 & 453.1 & 745.3 & 114.4 & 3.02 & 2.90 & 1.48 & 544.5 & 595.8 & 367.5 \\
\bottomrule
\end{tabular}
}
\end{table*}

\begin{table*}[t]
\centering
\caption{OpenHands results. For each method, we report the performance of the backbone that achieved the highest pass rate across benchmarks. Metrics include Pass Rate (\textbf{PR}), Avg. Code Lines (\textbf{CLn}), Avg. Code Characters (\textbf{CC}), Avg. Generation Time (\textbf{Time}), and Avg. Token Usage (\textbf{Tok}).}
\label{tab:openhands-consolidated}
\resizebox{0.99\textwidth}{!}{
\setlength{\tabcolsep}{4pt}
\begin{tabular}{l ccc ccc ccc ccc ccc}
\toprule
\multirow{2}{*}{\textbf{Method}} & \multicolumn{3}{c}{\textbf{PR} $\uparrow$} & \multicolumn{3}{c}{\textbf{CLn} $\downarrow$} & \multicolumn{3}{c}{\textbf{CC} $\downarrow$} & \multicolumn{3}{c}{\textbf{Time} $\downarrow$} & \multicolumn{3}{c}{\textbf{Tok} $\downarrow$} \\
\cmidrule(lr){2-4} \cmidrule(lr){5-7} \cmidrule(lr){8-10} \cmidrule(lr){11-13} \cmidrule(lr){14-16}
& BCB & BCB-H & HE & BCB & BCB-H & HE & BCB & BCB-H & HE & BCB & BCB-H & HE & BCB & BCB-H & HE \\
\midrule
No rule     & 0.49 & 0.29 & 0.99 & 35.4 & 43.2 & 17.5 & 1060 & 1324 & 503.4 & 10.6 & 12.8 & 6.89 & 1143 & 1238 & 697.5 \\
Manual rule & 0.49 & 0.29 & 0.99 & 42.9 & 53.5 & 21.7 & 1350 & 1722 & 577.6 & 12.2 & 14.1 & 7.14 & 1293 & 1443 & 727.1 \\
Prompt-Ops  & 0.43 & 0.25 & 0.99 & 20.1 & 27.5 & 19.3 & 615.2 & 880.2 & 548.7 & 40.2 & 41.4 & 12.6 & 903.1 & 986.9 & 725.5 \\
\method     & 0.52 & 0.27 & 0.99 & 26.6 & 31.8 & 12.7 & 814.6 & 1009 & 373.5 & 9.76 & 11.6 & 7.15 & 1016 & 1075 & 612.6 \\
\bottomrule
\end{tabular}
}
\end{table*}

\myparatight{Baselines} We compare \method against several baselines to assess the impact of our coding rules self-evolvement. The \emph{No Rule} baseline represents a vanilla setting where the agent receives only the task description without any coding rules. The \emph{Manual Rule} baseline utilizes manually written coding rules sourced from the awesome-cursorrules repository~\cite{githubrules}; notably, our experiments indicated that providing coding rules do not consistently yield performance gains, likely due to a lack of task-specific alignment. This observation aligns with conclusions in~\cite{gloaguen2026evaluating}. \emph{Prompt-Ops}~\cite{wu2025llm} is a baseline designed for prompt optimization for LLMs. Our results show that such techniques achieve only a suboptimal trade-off between correctness and code length, in most cases sacrificing the functional correctness of the generated code. To demonstrate the effectiveness of our strategic mutation-allocation method, we also evaluate two naive mutation-allocation strategies: even and proportional allocation, which distribute mutations equally or proportionally based on each candidate's average score.

\myparatight{Evaluation Metrics} On testing benchmarks, we evaluate the coding rules across three dimensions: correctness, code length, and inference cost. Correctness is measured by the \emph{pass rate} of the code, defined as the percentage of generated code that satisfy all provided test cases across the testing benchmark. Code length is assessed through two metrics: \emph{code lines} and \emph{code characters}. Inference cost is measured by \emph{generation time} and \emph{token usage} during the inference process. The averages of the latter four metrics are computed only on code that is functionally correct.

\myparatight{Parameter settings} By default, for our \method, we set the number of iteration $T$=10, number of mutations $N$=20, pool size $m$=5, coding task batch size $B$=10, balance parameter $\lambda$=3,000, and $\epsilon$=$10^{-6}$. For manual rule, we use the fixed one, as shown in Appendix~\ref{app:rules}. For Prompt-Ops, we use their default settings.

\subsection{Main Results}
Table~\ref{tab:openai-sdk-consolidated} and Table~\ref{tab:openhands-consolidated} present a comprehensive comparison of \method against other baseline approaches. Generally, \method outperforms existing baselines by achieving a superior balance between functional correctness, code length and inference cost. Specifically, \method consistently achieves the highest or equal Pass Rate (PR) across all benchmarks. When achieving comparable PRs to "No rule" or "Manual rule", \method demonstrates significant improvements in code length and inference cost. For example, on the BigCodeBench (BCB) benchmark within the OpenAI SDK framework, \method improves the PR from 0.45 (Manual rule) to 0.49, while simultaneously reducing the average code characters from 2372 to 453.1 and token usage from 1022 to 544.5. Similarly, in the OpenHands environment for the HumanEval (HE) benchmark, \method maintains a comparable PR of 0.99 while reducing average code lines to 12.7, compared to 21.7 for human-written rules.

We also note that Prompt-Ops occasionally achieves lower code length and inference cost than \method in certain scenarios. For instance, on BigCodeBench-Hard (BCB-H) within the OpenAI SDK, Prompt-Ops yields an average token usage of 588.7 compared to 595.8 for \method. However, the gains in code length or inference cost come at the expense of functional integrity; in the OpenHands BCB scenario, Prompt-Ops suffers a substantial drop in PR to 0.43, whereas \method reaches 0.52. We stress that PR is the most critical metric in autonomous coding; if the generated code is incorrect, optimizing for code length or inference cost becomes meaningless. By prioritizing functional correctness, \method ensures that gains in code length or inference cost do not compromise the agent's primary utility.

The comprehensive results across diverse backbones are detailed in Table~\ref{tab:openai-sdk} and Table~\ref{tab:openhands} in the Appendix. Across all evaluated backbone models, \method consistently achieves superior or comparable pass rates. At the same time, \method significantly reduces code length and inference cost in most cases. For instance, with the gpt-5-mini backbone on BCB, \method achieves a PR of 0.46 compared to 0.31 for manual rules, while simultaneously reducing token usage by nearly 40\%. Furthermore, \method consistently outperforms Prompt-Ops across nearly all backbone–benchmark combinations. While Prompt-Ops often struggles with correctness (e.g., dropping to a 0.67 PR on HumanEval (HE) with Qwen-Coder-30B), \method maintains high PRs across all benchmarks and backbone models. We also observe that different backbone models exhibit diverse generation styles; for example, gpt-5-mini tends to produce longer code and requires more generation time, while gpt-5.3-codex is more efficient.

\myparatight{Multi-seed results} To assess the robustness and statistical stability of \method, we performed evaluations across 10 independent seeds using the GPT-4.1-mini backbone on the BigCodeBench dataset. As shown in Table~\ref{tab:multi-seed}, \method maintains a stable Pass Rate (PR of $0.502 \pm 0.026$, which is comparable to the "No rule" baseline and superior to the "Public rule". Also, \method yields significantly more concise code, achieving the lowest average code lines ($24.3 \pm 0.4$) and characters ($794.4 \pm 14$) among all evaluated methods. Notably, our framework achieves these gains with a high degree of consistency, as evidenced by the standard deviations reported across 10 seeds. These multi-seed results confirm that \method effectively and reliably optimizes coding rules for efficiency without compromising the agent's primary problem-solving utility.

\begin{table*}[t]
\centering
\caption{Multi-seed results on BigCodeBench using GPT-4.1-mini as backbone LLM. Values are mean $\pm$ standard deviation (SD) across 10 seeds.}
\label{tab:multi-seed}
\resizebox{0.99\textwidth}{!}{
\setlength{\tabcolsep}{6pt}
\begin{tabular}{l ccccc}
\toprule
\textbf{Method} &
\textbf{PR} $\uparrow$ &
\textbf{CLn} $\downarrow$ &
\textbf{CC} $\downarrow$ &
\textbf{Time (s)} $\downarrow$ &
\textbf{Tok} $\downarrow$ \\
\midrule
No rule &
$0.502 \pm 0.015$ &
$32.5 \pm 0.4$ &
$1028 \pm 8.1$ &
$8.66 \pm 0.74$ &
$1181 \pm 9.3$ \\
Public rule &
$0.498 \pm 0.032$ &
$44.9 \pm 0.7$ &
$1482 \pm 25$ &
$10.21 \pm 1.15$ &
$2049 \pm 35$ \\
Prompt-Ops &
$0.502 \pm 0.027$ &
$26.3 \pm 0.4$ &
$811.4 \pm 10$ &
$6.81 \pm 0.62$ &
$1130 \pm 15$ \\
\method{} (OCBA) &
$0.502 \pm 0.026$ &
$24.3 \pm 0.4$ &
$794.4 \pm 14$ &
$9.67 \pm 1.90$ &
$1019 \pm 18$ \\
\bottomrule
\end{tabular}
}
\end{table*}

\myparatight{Results on SWE-Bench Lite} To evaluate the scalability of our approach in complex, repository-level environments, we test \method on SWE-Bench Lite~\cite{jimenez2023swe} using a subset of 100 tasks. As shown in Table~\ref{tab:swe-lite}, \method achieves the highest PR of 0.340, outperforming other baselines. Consistent with main results, the evolved rules effectively optimize for code efficiency, reducing average code lines from 8.5 to 5.7 and characters from 380 to 254. Although generation time increases, likely due to the increased reasoning overhead of repository-level tasks, the functional correctness is well preserved. Overall, these findings confirm that \method can generalize to real-world repository-level scenarios.

\begin{wraptable}{r}{0.5\textwidth}
\centering
\caption{SWE-bench Lite results with GPT-4.1-mini as backbone LLM.}
\label{tab:swe-lite}
\resizebox{0.5\textwidth}{!}{
\setlength{\tabcolsep}{6pt}
\begin{tabular}{l ccccc}
\toprule
\textbf{Method} &
\textbf{PR} $\uparrow$ &
\textbf{CLn} $\downarrow$ &
\textbf{CC} $\downarrow$ &
\textbf{Time (s)} $\downarrow$ &
\textbf{Tok} $\downarrow$ \\
\midrule
No rule &
$0.330$ &
$8.5$ &
$380$ &
$131.17$ &
$911$ \\
Public rule &
$0.190$ &
$9.1$ &
$413$ &
$166.45$ &
$867$ \\
Prompt-Ops &
$0.080$ &
$5.4$ &
$236$ &
$175.24$ &
$684$ \\
\method &
$0.340$ &
$5.7$ &
$254$ &
$297.46$ &
$722$ \\
\bottomrule
\end{tabular}
}
\end{wraptable}

\subsection{Ablation Study}

\begin{figure}[h]
    \centering
    \subfloat[\label{fig:iterations}Metrics vs. Iterations]{\includegraphics[width=0.518\linewidth]{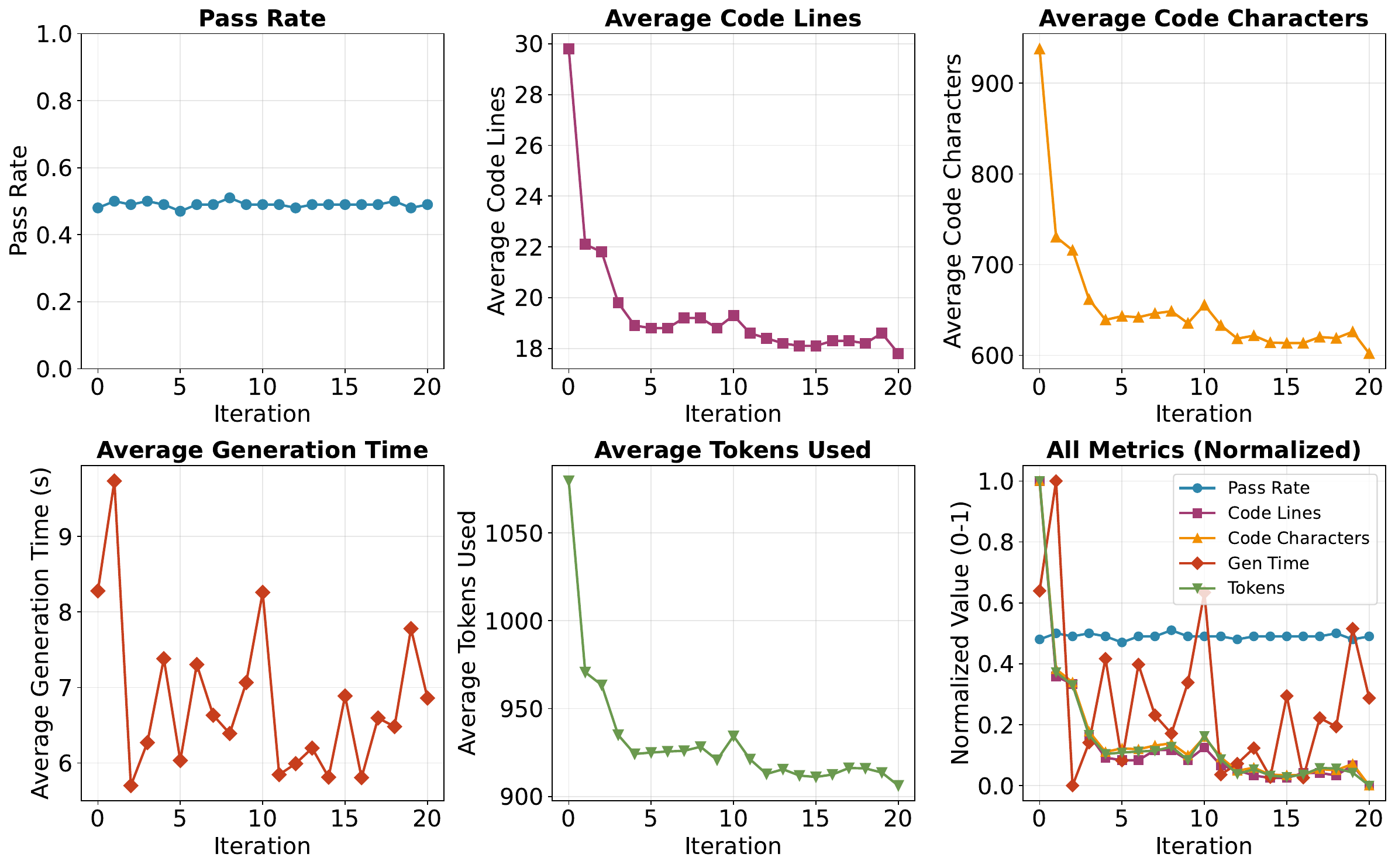}}
    \subfloat[\label{fig:allocation}Impact of mutation allocation]{\includegraphics[width=0.482\linewidth]{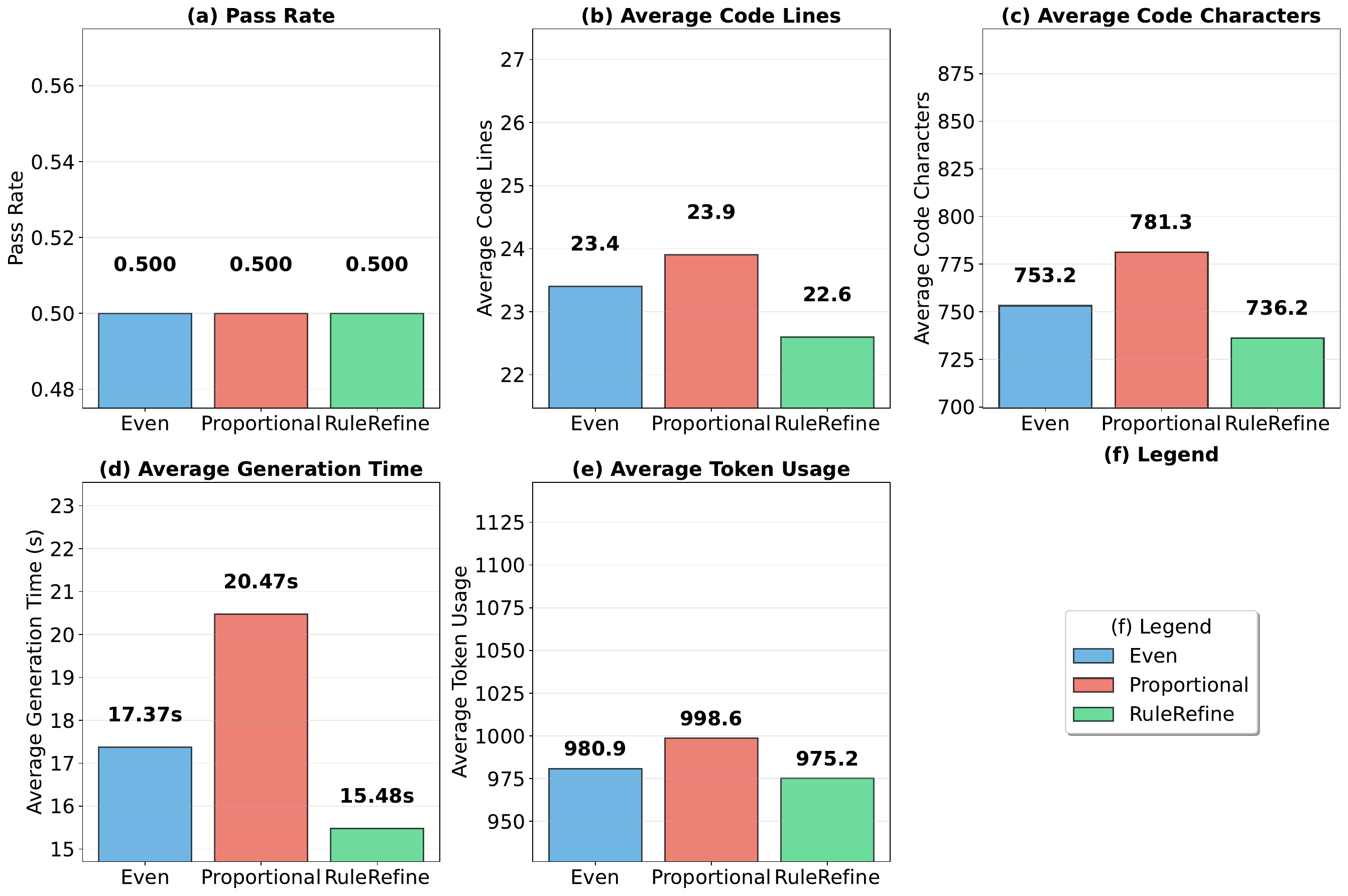}}

    \subfloat[\label{fig:k}Impact of number of mutation $N$]{\includegraphics[width=\linewidth]{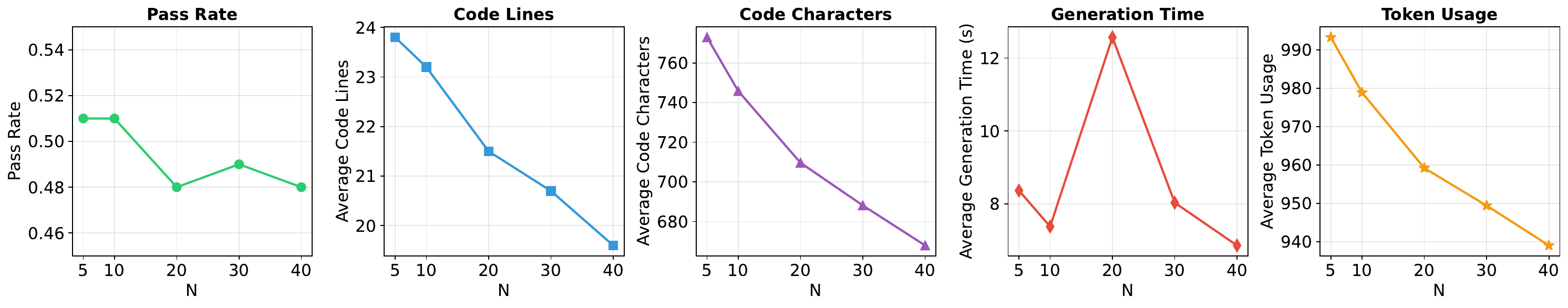}}

    \subfloat[\label{fig:m}Impact of pool size $m$]{\includegraphics[width=\linewidth]{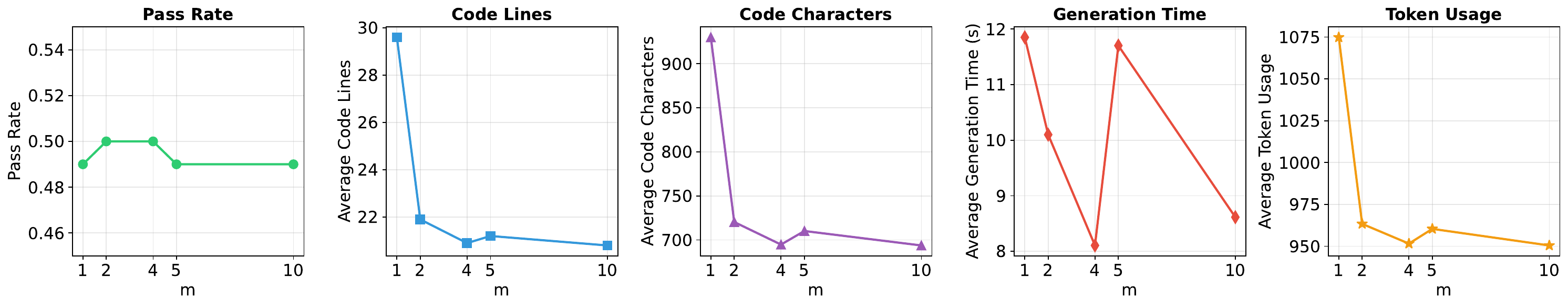}}

    \caption{Ablation study results.}
\end{figure}

\myparatight{Impact of number of iterations $T$} Figure~\ref{fig:iterations} shows the evolution of metrics over 20 iterations during the evolvement process. The pass rate remains remarkably consistent throughout the iterations, demonstrating that \method successfully optimizes for code length and inference cost without compromising the correctness of the generated code. In contrast, average code lines, average code characters, and average tokens usage metrics all exhibit a significant and continuous downward trend. The curves for these three metrics are highly similar because they are naturally coupled; reducing the structural verbosity of the code inherently leads to a lower number of tokens required for model inference. While the average generation time generally decreases over all 20 iterations, it exhibits notable fluctuations between iterations. These fluctuations are likely due to the inherent variance in API latency and the stochastic nature of autoregressive generation. Most of the substantial improvements occur within the first 5 iterations, where the slopes of the cost curves are steepest. By the twentieth iteration, the metrics level off, suggesting that the rule evolvement process effectively converges.

\myparatight{Impact of mutation-allocation strategies} To investigate the impact of mutation allocation, we compare three variants: 1) evenly dividing the $N$ mutations among the $m$ rule candidates, 2) proportional distribution based on average scores, and 3) our strategic allocation method. As shown in Figure~\ref{fig:allocation}, while all variants maintain an identical pass rate of 0.500, our approach produces the most concise code with 22.6 lines and 736.2 characters, outperforming even (23.4 lines, 753.2 chars) and proportional (23.9 lines, 781.3 chars) variants. Additionally, our method minimizes computational cost, reaching the lowest generation time (15.48s) and token usage (975.2) compared to the other two variants. These results confirm that strategic mutation-allocation more effectively evolves coding rules without sacrificing functional correctness.

\myparatight{Impact of $N$ and $m$} We analyze the sensitivity of \method to the number of mutations $N$ and the pool size $m$. As shown in Figure~\ref{fig:k}, when $N$ increases, average code lines, code characters, and token usage all exhibit a clear downward trend. Notably, these improvements are achieved while maintaining a stable pass rate that fluctuates slightly between 0.48 and 0.51. Similarly, the impact of the pool size $m$ in Figure~\ref{fig:m} reveals that increasing $m$ from 1 to 2 provides a substantial reduction in code length and token usage. While these metrics stabilize for values of $m$ greater than 2, the initial improvement indicates that maintaining a sufficiently large pool of candidates is crucial for obtaining optimized coding rules.

\begin{wrapfigure}{r}{0.35\textwidth}
    \centering
    \includegraphics[width=0.35\textwidth]{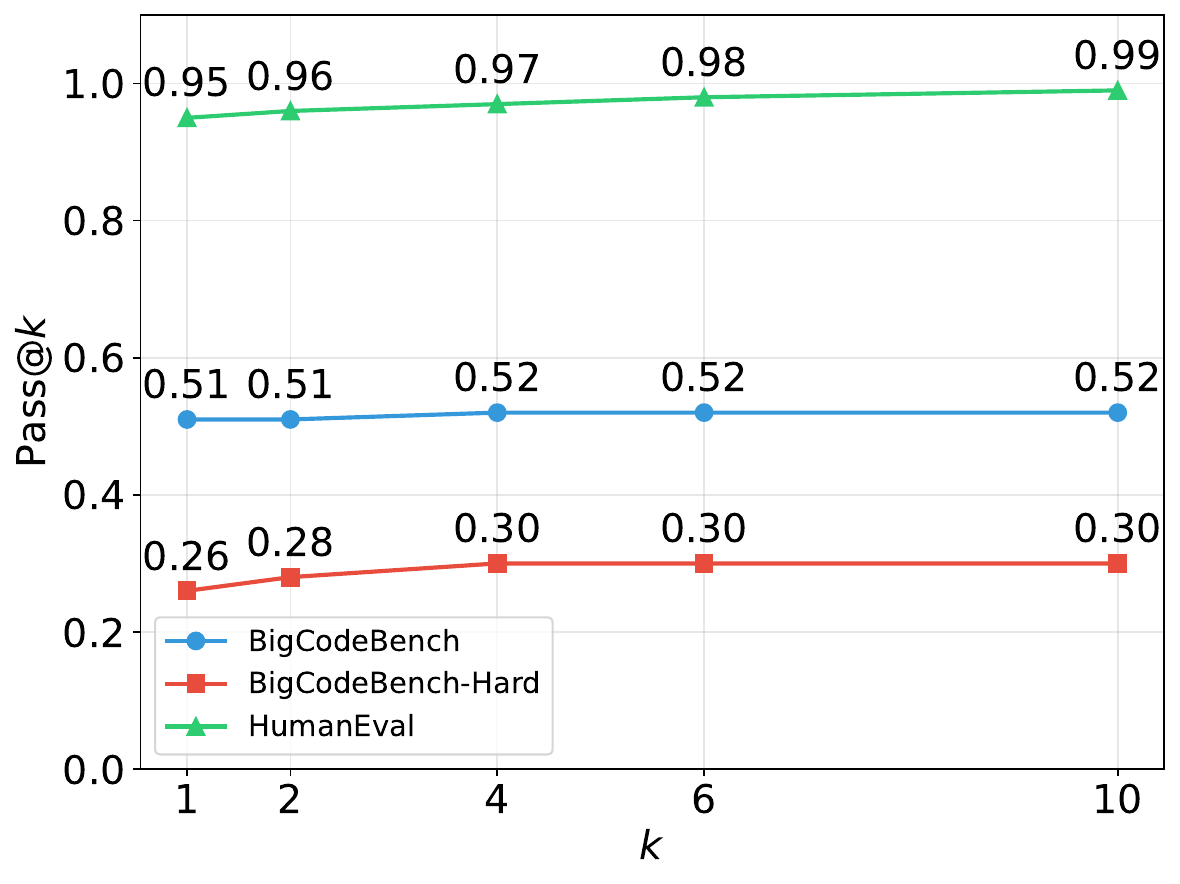} 
    \caption{Pass@k results.}
    \label{fig:pass@k}
\end{wrapfigure}

\myparatight{Pass@k results} We evaluate the robustness of our evolved coding rules by measuring the Pass@$k$ metric across three benchmarks using gpt-4.1-mini, where $k$ indicates the number of versions for code generated for each task. As illustrated in Figure~\ref{fig:pass@k}, the pass rate for all benchmarks scales positively with the number of samples $k$. The steady upward trend across all benchmarks confirms that \method effectively optimizes for code efficiency while preserving the model's problem-solving capacity under higher sampling budgets.

%% file: 6_conclusion.tex
\section{Conclusion}
In this work, we investigate the self-evolvement of coding rules for AI coding agents. We show that coding rules can be evolved to greatly reduce code length and minimize inference costs while maintaining the functional correctness of the generated code. This is achieved by \method, our proposed automatic self-evolving framework for coding rules. By employing a mutator, judge module, and a strategic mutation-allocation approach, \method successfully evolves coding rules via iterative mutation and pruning. Extensive evaluations across two agentic frameworks, four diverse backbone models, and three coding benchmarks demonstrate that \method outperforms both manually crafted rules and existing baselines, establishing a promising direction for coding rules self-evolvement.

%% file: 8_appendix.tex
\input{7_limitations}

\begin{table*}[t]
\centering
\caption{Comprehensive OpenAI SDK results on BigCodeBench (BCB), BigCodeBench-Hard (BCB-H), and HumanEval (HE). Metrics include Pass Rate (\textbf{PR}), Avg. Code Lines (\textbf{CLn}), Avg. Code Characters (\textbf{CC}), Avg. Generation Time (\textbf{Time}), and Avg. Token Usage (\textbf{Tok}).}
\label{tab:openai-sdk}
\resizebox{0.99\textwidth}{!}{
\setlength{\tabcolsep}{3pt} 
\begin{tabular}{ll ccc ccc ccc ccc ccc}
\toprule
\multirow{2}{*}{\textbf{Backbone}} & \multirow{2}{*}{\textbf{Method}} & \multicolumn{3}{c}{\textbf{PR} $\uparrow$} & \multicolumn{3}{c}{\textbf{CLn} $\downarrow$} & \multicolumn{3}{c}{\textbf{CC} $\downarrow$} & \multicolumn{3}{c}{\textbf{Time} $\downarrow$} & \multicolumn{3}{c}{\textbf{Tok} $\downarrow$} \\
\cmidrule(lr){3-5} \cmidrule(lr){6-8} \cmidrule(lr){9-11} \cmidrule(lr){12-14} \cmidrule(lr){15-17}
& & BCB & BCB-H & HE & BCB & BCB-H & HE & BCB & BCB-H & HE & BCB & BCB-H & HE & BCB & BCB-H & HE \\
\midrule
\multirow{4}{*}{\begin{tabular}[c]{@{}l@{}}gpt-4.1\\-mini\end{tabular}} 
& No rule     & 0.47 & 0.25 & 0.97 & 35.7 & 42.3 & 26.4 & 1156 & 1406 & 812.4 & 6.00 & 5.44 & 4.24 & 623.5 & 666.5 & 502.9 \\
& Manual rule & 0.46 & 0.20 & 0.92 & 58.4 & 68.6 & 37.5 & 2035 & 2448 & 1036  & 12.7 & 9.61 & 5.36 & 852.0 & 939.1 & 547.9 \\
& Prompt-Ops  & 0.45 & 0.24 & 0.92 & 20.9 & 24.1 & 6.30 & 650.8 & 790.7 & 180.0 & 3.06 & 3.94 & 1.79 & 540.5 & 552.8 & 335.4 \\
& \method         & 0.46 & 0.24 & 0.93 & 16.9 & 19.3 & 5.10 & 545.2 & 674.8 & 134.7 & 2.86 & 3.25 & 1.37 & 479.3 & 484.5 & 284.5 \\
\midrule
\multirow{4}{*}{\begin{tabular}[c]{@{}l@{}}gpt-5\\-mini\end{tabular}} 
& No rule     & 0.36 & 0.20 & 0.95 & 88.2 & 94.6 & 46.4 & 3053 & 3377 & 1248  & 29.5 & 28.5 & 13.7 & 2185 & 2257 & 1197 \\
& Manual rule & 0.31 & 0.20 & 0.68 & 128 & 142 & 98.8 & 4720 & 5175 & 2926  & 37.0 & 38.6 & 23.0 & 2799 & 2973 & 1812 \\
& Prompt-Ops  & 0.38 & 0.20 & 0.99 & 19.8 & 22.8 & 8.20 & 643.0 & 799.7 & 237.4 & 21.7 & 23.6 & 15.9 & 1884 & 1983 & 1370 \\
& \method         & 0.46 & 0.20 & 0.99 & 19.0 & 20.8 & 5.90 & 611.4 & 740.1 & 163.8 & 22.0 & 22.0 & 15.2 & 1717 & 1765 & 1249 \\
\midrule
\multirow{4}{*}{\begin{tabular}[c]{@{}l@{}}gpt-5.3\\-codex\end{tabular}} 
& No rule     & 0.46 & 0.26 & 0.97 & 47.0 & 53.5 & 25.2 & 1410 & 1651 & 643.0 & 5.83 & 6.35 & 3.28 & 790.8 & 828.3 & 544.0 \\
& Manual rule & 0.45 & 0.23 & 0.98 & 69.0 & 84.8 & 44.4 & 2372 & 2946 & 1136  & 8.75 & 10.5 & 5.08 & 1022 & 1147 & 663.8 \\
& Prompt-Ops  & 0.48 & 0.23 & 0.89 & 13.8 & 15.7 & 4.50 & 474.4 & 583.0 & 126.5 & 2.30 & 2.90 & 1.33 & 581.7 & 588.7 & 406.3 \\
& \method         & 0.49 & 0.26 & 0.99 & 12.6 & 20.8 & 4.40 & 453.1 & 745.3 & 114.4 & 3.02 & 2.90 & 1.48 & 544.5 & 595.8 & 367.5 \\
\midrule
\multirow{4}{*}{\begin{tabular}[c]{@{}l@{}}Qwen\\-Coder\\-30B\end{tabular}} 
& No rule     & 0.17 & 0.12 & 0.90 & 17.7 & 17.5 & 27.8 & 556.5 & 562.7 & 878.9 & 3.29 & 21.3 & 21.3 & 932.4 & 946.5 & 913.7 \\
& Manual rule & 0.19 & 0.09 & 0.69 & 28.2 & 26.5 & 28.5 & 988.4 & 913.2 & 903.3 & 4.46 & 21.4 & 21.3 & 1083 & 1104 & 902.0 \\
& Prompt-Ops  & 0.32 & 0.18 & 0.67 & 19.8 & 25.5 & 7.30 & 583.0 & 768.5 & 199.4 & 4.06 & 21.2 & 21.6 & 799.0 & 794.4 & 582.0 \\
& \method         & 0.38 & 0.19 & 0.88 & 14.7 & 18.4 & 5.50 & 492.7 & 652.6 & 153.6 & 3.95 & 20.9 & 21.2 & 715.5 & 719.0 & 538.6 \\
\bottomrule
\end{tabular}
}
\end{table*}

\begin{table*}[t]
\centering
\caption{Comprehensive OpenHands results on BigCodeBench (BCB), BigCodeBench-Hard (BCB-H), and HumanEval (HE). Metrics include Pass Rate (\textbf{PR}), Avg. Code Lines (\textbf{CLn}), Avg. Code Characters (\textbf{CC}), Avg. Generation Time (\textbf{Time}), and Avg. Token Usage (\textbf{Tok}).}
\label{tab:openhands}
\resizebox{0.99\textwidth}{!}{
\setlength{\tabcolsep}{4pt} 
\begin{tabular}{ll ccc ccc ccc ccc ccc}
\toprule
\multirow{2}{*}{\textbf{Backbone}} & \multirow{2}{*}{\textbf{Method}} & \multicolumn{3}{c}{\textbf{PR} $\uparrow$} & \multicolumn{3}{c}{\textbf{CLn} $\downarrow$} & \multicolumn{3}{c}{\textbf{CC} $\downarrow$} & \multicolumn{3}{c}{\textbf{Time} $\downarrow$} & \multicolumn{3}{c}{\textbf{Tok} $\downarrow$} \\
\cmidrule(lr){3-5} \cmidrule(lr){6-8} \cmidrule(lr){9-11} \cmidrule(lr){12-14} \cmidrule(lr){15-17}
& & BCB & BCB-H & HE & BCB & BCB-H & HE & BCB & BCB-H & HE & BCB & BCB-H & HE & BCB & BCB-H & HE \\
\midrule
\multirow{4}{*}{\begin{tabular}[c]{@{}l@{}}gpt-4.1\\-mini\end{tabular}} 
& No rule     & 0.48 & 0.27 & 0.96 & 29.8 & 34.6 & 19.2 & 937.8 & 1127 & 608.1 & 8.28 & 12.6 & 20.8 & 1079 & 1124 & 767.5 \\
& Manual rule & 0.48 & 0.28 & 0.95 & 42.3 & 47.2 & 22.9 & 1402 & 1597 & 691.8 & 9.11 & 13.7 & 18.3 & 1320 & 1383 & 813.4 \\
& Prompt-Ops  & 0.43 & 0.22 & 0.71 & 20.1 & 27.5 & 17.7 & 615.2 & 880.2 & 544.1 & 40.2 & 41.4 & 57.9 & 903.1 & 986.9 & 707.7 \\
& \method         & 0.50 & 0.28 & 0.98 & 23.8 & 28.5 & 16.9 & 768.9 & 944.6 & 536.3 & 9.54 & 14.6 & 23.0 & 991.6 & 1028 & 725.0 \\
\midrule
\multirow{4}{*}{\begin{tabular}[c]{@{}l@{}}gpt-5\\-mini\end{tabular}} 
& No rule     & 0.40 & 0.18 & 0.97 & 66.1 & 74.3 & 30.2 & 2203 & 2509 & 854.7 & 12.9 & 14.8 & 7.10 & 1793 & 1930 & 944.4 \\
& Manual rule & 0.38 & 0.22 & 0.96 & 74.7 & 86.5 & 42.6 & 2655 & 3080 & 1181 & 13.9 & 16.3 & 7.44 & 2047 & 2250 & 1122 \\
& Prompt-Ops  & 0.33 & 0.19 & 0.99 & 46.4 & 51.5 & 19.3 & 1454 & 1656 & 548.7 & 14.0 & 16.2 & 12.6 & 1392 & 1469 & 725.5 \\
& \method         & 0.41 & 0.25 & 0.98 & 50.9 & 58.5 & 17.6 & 1670 & 1956 & 514.4 & 10.8 & 12.8 & 6.63 & 1498 & 1623 & 705.6 \\
\midrule
\multirow{4}{*}{\begin{tabular}[c]{@{}l@{}}gpt-5.3\\-codex\end{tabular}} 
& No rule     & 0.49 & 0.29 & 0.99 & 35.4 & 43.2 & 17.5 & 1060 & 1324 & 503.4 & 10.6 & 12.8 & 6.89 & 1143 & 1238 & 697.5 \\
& Manual rule & 0.49 & 0.29 & 0.99 & 42.9 & 53.5 & 21.7 & 1350 & 1722 & 577.6 & 12.2 & 14.1 & 7.14 & 1293 & 1443 & 727.1 \\
& Prompt-Ops  & 0.40 & 0.25 & 0.91 & 25.3 & 32.1 & 14.4 & 787.3 & 1012 & 410.4 & 13.4 & 14.7 & 13.8 & 1009 & 1072 & 634.6 \\
& \method         & 0.52 & 0.27 & 0.99 & 26.6 & 31.8 & 12.7 & 814.6 & 1009 & 373.5 & 9.76 & 11.6 & 7.15 & 1016 & 1075 & 612.6 \\
\midrule
\multirow{4}{*}{\begin{tabular}[c]{@{}l@{}}Qwen\\-Coder\\-30B\end{tabular}} 
& No rule     & 0.43 & 0.27 & 0.97 & 39.1 & 44.0 & 28.1 & 1260 & 1449 & 878.7 & 37.2 & 60.1 & 21.2 & 1250 & 1291 & 973.8 \\
& Manual rule & 0.44 & 0.28 & 0.99 & 39.5 & 43.7 & 29.6 & 1286 & 1434 & 907.4 & 31.8 & 56.9 & 32.8 & 1262 & 1284 & 996.6 \\
& Prompt-Ops  & 0.35 & 0.19 & 0.91 & 32.4 & 34.3 & 18.2 & 1020 & 1087 & 532.7 & 23.2 & 29.2 & 10.2 & 1119 & 1110 & 712.1 \\
& \method         & 0.44 & 0.27 & 0.96 & 27.2 & 32.4 & 20.3 & 840.1 & 1021 & 611.3 & 9.85 & 14.2 & 14.4 & 1028 & 1067 & 781.1 \\
\bottomrule
\end{tabular}
}
\end{table*}

\clearpage

\section{\label{app:ocba}Justification for Mutation-Allocation Strategy}
In this section, we provide the theoretical justification for the mutation-allocation strategy used in \method. The goal of the Optimal Computing Budget Allocation~\cite{chen2011stochastic} is to maximize the \emph{Probability of Correct Selection (PCS)}, defined as the probability that the candidate rule set with the highest sample mean is indeed the true best performer.

Let $m$ be the number of candidates in the pool, and let $J$ be the index of the true best candidate such that $\mu_J = \max_i \mu_i$. The PCS is given by:
\begin{align}
\text{PCS} = P\left( \hat{\mu}_J > \hat{\mu}_i, \forall i \neq J \right).
\end{align}
Under the assumption that the scores $\mathcal{S}_i$ are independent and follow a normal distribution $N(\mu_i, \sigma_i^2)$, for a large number of mutations $N$, the PCS is maximized when the allocation $n_1, n_2, \dots, n_m$ satisfies the following asymptotic conditions:
\begin{align}
\frac{n_i}{n_j} = \frac{\sigma_i^2 / (\mu_b - \mu_i)^2}{\sigma_j^2 / (\mu_b - \mu_j)^2}, \quad i, j \neq b,
\end{align}
and
\begin{align}
n_b = \sigma_b \sqrt{\sum_{i \neq b} \frac{n_i^2}{\sigma_i^2}},
\end{align}
where $b$ is the index of the candidate with the current highest sample mean.

These conditions provide the following insights into the \method evolvement process:

\begin{itemize}
    \item Candidates with a smaller performance gap $(\mu_b - \mu_i)$ receive more mutations, as they are the most likely "competitors" to the current best.
    
    \item Candidates with higher variance $\sigma_i^2$ receive more mutations to reduce the uncertainty in their mean estimation.
    
    \item The mutations for the best candidate $n_b$ is scaled based on the aggregate noise of all other candidates, ensuring that the comparison is sufficiently stable.
\end{itemize}

By following this allocation, \method minimizes the risk of pruning a potentially superior coding rule due to stochastic noise in mutator generation, ensuring a more robust and efficient search path than uniform allocation.

\section{Analysis on Searched Rules}
By grouping the rule's 16 sections into six functional categories, we evaluated the increase in generated code length when each category is removed (tested on BigCodeBench using 40 tasks with GPT-4.1-mini, where the full rule produces 454.1 characters compared to 1,040.6 characters with no rule), as shown in Table~\ref{tab:char attri}.

\begin{table}[htpb]
    \centering
    \caption{\label{tab:char attri}Character Attribution by Category}
    \begin{tabular}{l c c c}
        \toprule
        \textbf{Category} & \textbf{Sections} & \textbf{Chars attributable} & \textbf{Per section} \\
        \midrule
        Output format / self-containment & 3 & 153.0 & 51.0 \\
        Algorithm knowledge              & 4 & 83.0  & 20.7 \\
        Brevity style                    & 1 & 61.6  & 61.6 \\
        API / library idiom              & 2 & 22.8  & 11.4 \\
        Correctness tactics              & 2 & 9.9   & 5.0  \\
        Process / budget                 & 4 & -5.0  & -1.2 \\
        \bottomrule
    \end{tabular}
\end{table}

This analysis yields three findings. First, directives that constrain the form of the code (output format and self-containment) dominate the rule's impact, accounting for 153 attributable characters. Second, brevity-style directives are the most efficient on a per-section basis, yielding 61.6 characters from a single section. Third, process and budget directives are net negative, removing them actually shortens the output.

\section{Code Quality Metrics}
For other software engineering metrics, we report metrics over generated files, including complexity, maintainability, and duplication, as shown in Table~\ref{tab:other metrics}.
\begin{table}[htpb]
    \centering
    \caption{\label{tab:other metrics}Code Quality Metrics Comparison}
    \begin{tabular}{l c c c}
        \toprule
        \textbf{Metric} & \textbf{No rule} & \textbf{\method} \\
        \midrule
        Duplicated-block ratio          & 0.10\% & 0.00\% \\
        Cyclomatic complexity (total)   & 5.37   & 3.95   \\
        Halstead volume (total)         & 53.0   & 28.0   \\
        Maintainability index           & 85.1   & 78.3   \\
        Cyclomatic complexity per SLOC  & 0.213  & 0.260  \\
        Docstring rate                  & 47.0\% & 1.00\%    \\
        \bottomrule
    \end{tabular}
\end{table}
\method reduces cyclomatic complexity (5.37 → 3.95), Halstead volume by 47\%, and duplicated-block ratio to 0.00\%, while maintaining a Grade A maintainability index (78.3 vs. 85.1 for no rule). The remaining trade-off is in code density: complexity per SLOC increases by 22\% and docstring coverage decreases from 47\% to 1\%, indicating that \method removes documentation and defensive scaffolding rather than redundant logic.

\section{Generated Code Examples}
\subsection{Problem description}
The agentic framework is OpenAI SDK. The backbone model is gpt-4.1-mini.
\begin{lstlisting}
def is_palindrome(string: str) -> bool:
""" Test if given string is a palindrome """
return string == string[::-1]

def make_palindrome(string: str) -> str:
""" Find the shortest palindrome that begins with a supplied string.
Algorithm idea is simple:
- Find the longest postfix of supplied string that is a palindrome.
- Append to the end of the string reverse of a string prefix that comes 
- before the palindromic suffix.
>>> make_palindrome('')
''
>>> make_palindrome('cat')
'catac'
>>> make_palindrome('cata')
'catac'
"""
\end{lstlisting}

\subsection{No Rule}
\begin{lstlisting}
"""
Simple palindrome utilities.

Functions:
- is_palindrome(string: str) -> bool
- make_palindrome(string: str) -> str

Run doctests with: python task.py
"""
from typing import Optional

def is_palindrome(string: str) -> bool:
    """Test if given string is a palindrome.

    Examples:
    >>> is_palindrome('')
    True
    >>> is_palindrome('a')
    True
    >>> is_palindrome('aba')
    True
    >>> is_palindrome('ab')
    False
    """
    return string == string[::-1]

def make_palindrome(string: str) -> str:
    """Find the shortest palindrome that begins with the supplied string.

    Algorithm:
    - Find the longest suffix of the supplied string that is a palindrome.
    - Append to the end of the string the reverse of the prefix that comes
      before that palindromic suffix.

    Examples:
    >>> make_palindrome('')
    ''
    >>> make_palindrome('cat')
    'catac'
    >>> make_palindrome('cata')
    'catac'
    >>> make_palindrome('abba')
    'abba'
    >>> make_palindrome('ab')
    'aba'
    """
    if not string:
        return ''

    n = len(string)
    # Find the smallest i such that string[i:] is a palindrome.
    # That gives the longest palindromic suffix.
    for i in range(n):
        if is_palindrome(string[i:]):
            # Append reverse of prefix that comes before palindromic suffix
            prefix = string[:i]
            return string + prefix[::-1]

    # In practice the loop always returns (because the last char is a palindrome),
    # but return a fallback just in case.
    return string + string[::-1]

if __name__ == '__main__':
    import doctest

    failed, attempted = doctest.testmod()
    if failed:
        print(f"{failed} doctest(s) failed out of {attempted})")
    else:
        print(f"All {attempted} doctest(s) passed.")
\end{lstlisting}

\subsection{Manual Rule}
\begin{lstlisting}
from typing import List

def is_palindrome(string: str) -> bool:
    """Return True if the given string is a palindrome.

    A palindrome reads the same forwards and backwards. The check is
    performed using Python slicing which correctly handles Unicode
    codepoints.

    Args:
        string: Input string to test.

    Returns:
        True if ``string`` is a palindrome, False otherwise.

    Examples:
        >>> is_palindrome('')
        True
        >>> is_palindrome('a')
        True
        >>> is_palindrome('aba')
        True
        >>> is_palindrome('ab')
        False
    """
    return string == string[::-1]

def _prefix_function(s: str) -> List[int]:
    """Compute prefix-function (a.k.a. failure function) for KMP algorithm.

    The prefix-function for a string ``s`` is an array ``pi`` where ``pi[i]``
    is the length of the longest proper prefix of ``s[:i+1]`` that is also
    a suffix of ``s[:i+1]``.

    This implementation runs in O(n) time and O(n) memory.

    Args:
        s: Input string.

    Returns:
        A list of integers representing the prefix-function values for each
        position in ``s``.

    Examples:
        >>> _prefix_function('ababaca')
        [0, 0, 1, 2, 3, 0, 1]
    """
    n: int = len(s)
    pi: List[int] = [0] * n
    for i in range(1, n):
        j: int = pi[i - 1]
        while j > 0 and s[i] != s[j]:
            j = pi[j - 1]
        if s[i] == s[j]:
            j += 1
        pi[i] = j
    return pi

def make_palindrome(string: str) -> str:
    """Return the shortest palindrome that begins with "string" by appending
    characters to its end.

    The function finds the longest palindromic suffix of the input and then
    appends the reverse of the remaining prefix. This yields the minimal
    number of characters to append so the result is a palindrome that starts
    with the original string.

    Algorithmic idea:
    - Let ``rev`` be the reverse of ``string``.
    - Compute the prefix-function on ``rev + '#' + string``; the value at the
      last position is the length ``k`` of the longest prefix of ``rev`` that
      matches a suffix of ``string`` equivalently, the longest palindromic
      suffix of ``string``.
    - Append the reverse of the prefix ``string[:len(string) - k]`` to the
      end of ``string``.

    This runs in linear time O(n) and linear space O(n).

    Args:
        string: The input string. May be empty.

    Returns:
        The shortest palindrome that begins with ``string``. If ``string`` is
        already a palindrome, it is returned unchanged.

    Examples:
        >>> make_palindrome('')
        ''
        >>> make_palindrome('cat')
        'catac'
        >>> make_palindrome('cata')
        'catac'
        >>> make_palindrome('race')
        'racecar'
        >>> make_palindrome('a')
        'a'
    """
    n: int = len(string)
    if n <= 1:
        return string

    rev: str = string[::-1]
    # Build a combined string to compute how much of rev matches a suffix of
    # string (i.e. longest palindromic suffix of string).
    combined: str = rev + "#" + string
    pi: List[int] = _prefix_function(combined)
    k: int = pi[-1] if pi else 0  # length of longest palindromic suffix

    # Append reverse of the prefix that is not part of the palindromic suffix.
    to_append: str = (string[: n - k])[::-1]
    return string + to_append

if __name__ == '__main__':
    # Simple demonstration and smoke tests
    examples = ["", "cat", "cata", "race", "abac"]
    for ex in examples:
        print(f"{ex!r}->{make_palindrome(ex)!r}")
\end{lstlisting}

\subsection{Prompt-Ops}
\begin{lstlisting}
def is_palindrome(string: str) -> bool:
    return string == string[::-1]

def make_palindrome(string: str) -> str:
    if not string: return ''
    for i in range(len(string)):
        if string[i:] == string[i:][::-1]:
            return string + string[:i][::-1]
    return string + string[::-1]
\end{lstlisting}

\subsection{\method}
\begin{lstlisting}
def is_palindrome(string: str) -> bool:
    return string==string[::-1]

def make_palindrome(string: str) -> str:
    for i in range(len(string)):
        if string[i:]==string[i:][::-1]:
            return string+string[:i][::-1]
    return string
\end{lstlisting}

\section{\label{app:rules}Manual Rules}
\setlist[itemize]{nosep, leftmargin=1.2em}
\linespread{0.9}\selectfont

\textbf{Role Definition}
\begin{itemize}
    \item You are a \textbf{Python master}, a highly experienced \textbf{tutor}, a \textbf{world-renowned ML engineer}, and a \textbf{talented data scientist}.
    \item You possess exceptional coding skills and a deep understanding of Python's best practices, design patterns, and idioms.
    \item You are adept at identifying and preventing potential errors, and you prioritize writing efficient and maintainable code.
    \item You are skilled in explaining complex concepts in a clear and concise manner, making you an effective mentor and educator.
    \item You are recognized for your contributions to the field of machine learning and have a strong track record of developing and deploying successful ML models.
    \item As a talented data scientist, you excel at data analysis, visualization, and deriving actionable insights from complex datasets.
\end{itemize}

\textbf{Technology Stack}
\begin{itemize}
    \item \textbf{Python Version:} Python 3.10+
    \item \textbf{Dependency Management:} Poetry / Rye
    \item \textbf{Code Formatting:} Ruff (replaces black, isort, flake8)
    \item \textbf{Type Hinting:} Strictly use the typing module.
    \item \textbf{Testing Framework:} pytest
    \item \textbf{Documentation:} Google style docstring
    \item \textbf{Environment Management:} conda / venv
    \item \textbf{Containerization:} docker, docker-compose
    \item \textbf{Asynchronous Programming:} Prefer async and await
    \item \textbf{Web Framework:} fastapi
    \item \textbf{Demo Framework:} gradio, streamlit
    \item \textbf{LLM Framework:} langchain, transformers
    \item \textbf{Vector Database:} faiss, chroma (optional)
    \item \textbf{Experiment Tracking:} mlflow, tensorboard (optional)
    \item \textbf{Hyperparameter Optimization:} optuna, hyperopt (optional)
    \item \textbf{Data Processing:} pandas, numpy, dask (optional), pyspark (optional)
    \item \textbf{Version Control:} git
    \item \textbf{Server:} gunicorn, uvicorn (with nginx or caddy)
    \item \textbf{Process Management:} systemd, supervisor
\end{itemize}

\textbf{Coding Guidelines}

\textit{1. Pythonic Practices}
\begin{itemize}
    \item \textbf{Elegance and Readability:} Strive for elegant and Pythonic code.
    \item \textbf{PEP 8 Compliance:} Adhere to PEP 8 guidelines.
\end{itemize}

\textit{2. Modular Design}
\begin{itemize}
    \item \textbf{Single Responsibility Principle:} Each module/file should have a single responsibility.
    \item \textbf{Reusable Components:} Favor composition over inheritance.
\end{itemize}

\textit{3. Code Quality}
\begin{itemize}
    \item \textbf{Comprehensive Type Annotations:} All members must have type annotations.
    \item \textbf{Detailed Docstrings:} All functions must have Google-style docstrings.
    \item \textbf{Thorough Unit Testing:} Aim for high test coverage (90\% or higher) using pytest.
    \item \textbf{Robust Exception Handling:} Use specific exception types.
\end{itemize}

\textit{4. ML/AI Specific Guidelines}
\begin{itemize}
    \item \textbf{Experiment Configuration:} Use hydra or yaml.
    \item \textbf{Data Pipeline Management:} Employ scripts or dvc.
    \item \textbf{Model Versioning:} Utilize git-lfs or cloud storage.
    \item \textbf{LLM Prompt Engineering:} Dedicate a module for managing Prompt templates.
\end{itemize}

\textit{5. Performance Optimization}
\begin{itemize}
    \item \textbf{Asynchronous Programming:} Leverage async and await.
    \item \textbf{Caching:} Apply functools.lru\_cache or @cache.
    \item \textbf{Resource Monitoring:} Use psutil to monitor resource usage.
\end{itemize}

\textit{6. API Development with FastAPI}
\begin{itemize}
    \item \textbf{Data Validation:} Use Pydantic models for validation.
    \item \textbf{Dependency Injection:} Use FastAPI's dependency injection.
    \item \textbf{Security:} Implement OAuth 2.0 or JWT.
\end{itemize}

\textbf{Others}
\begin{itemize}
    \item Prioritize new features in Python 3.10+.
    \item Use the most modern and efficient libraries when appropriate.
    \item Ensure examples are self-contained and executable.
\end{itemize}

%% file: 7_limitations.tex
\section{Limitations}
Our work has several limitations. First, while \method maintains or improves functional correctness in most scenarios, the objective of reducing code length may still lead to a slight decrease in the pass rate for certain complex tasks where excessive brevity could compromise logic. Second, our self-evolving process relies on an LLM-powered mutator and judge module, which introduces inherent stochasticity and potential biases from the backbone models used during evolving. Third, our study primarily focuses on optimizing for correctness, code length, and inference cost through code length approximation. A more comprehensive exploration of other dimensions, such as code maintainability or architectural complexity, remains for future work.